\documentclass[sigconf,nonacm,balance=false]{acmart}
\usepackage{amsmath,booktabs,tikz,placeins,float,array,colortbl,tabularx}
\usepackage[ruled,vlined,linesnumbered]{algorithm2e}
\definecolor{rtrow}{RGB}{246,235,238}
\usetikzlibrary{arrows.meta,positioning,fit}
\setcopyright{none}
\acmConference{}{}{}
\acmDOI{}
\acmISBN{}
\renewcommand{\footnotetextcopyrightpermission}[1]{}

\newcommand{\cat}{\mathbin\Vert}
\title{InterFuse: Interaction-Guided Adaptive Fusion of Heterogeneous Models for Payment Risk Ranking}
\author{Haoyu Jiang}
\affiliation{\institution{Ant International}\country{}}
\email{wuxian.jhy@ant-intl.com}
\author{Weiyu Han}
\affiliation{\institution{Ant International}\country{}}
\author{Ze Zhao}
\affiliation{\institution{Ant International}\country{}}
\affiliation{\institution{Shanghai Jiao Tong University}\country{}}
\author{Jie Sheng}
\affiliation{\institution{Ant International}\country{}}
\author{Ruobing Zuo}
\affiliation{\institution{Ant International}\country{}}
\affiliation{\institution{East China Normal University}\country{}}
\author{Xiao Tan}
\affiliation{\institution{Ant International}\country{}}
\author{Shuai Chen}
\affiliation{\institution{Ant International}\country{}}
\author{Tianyi Zhang}
\affiliation{\institution{Ant International}\country{}}
\renewcommand{\shortauthors}{Haoyu Jiang et al.}
\hypersetup{pdfauthor={Haoyu Jiang, Weiyu Han, Ze Zhao, Jie Sheng, Ruobing Zuo, Xiao Tan, Shuai Chen, Tianyi Zhang}}
\begin{document}
\begin{abstract}
Payment platforms use view-specific models trained on transaction attributes, behavioral histories, and relational structures to rank transactions under limited review budgets. However, existing fusion methods often focus on prediction scores or interactions among model representations, with less attention to how these interactions should guide aggregation using each model's original prediction and input coverage. To address this limitation, we propose InterFuse, a fusion framework that uses a shared fusion state from cross-model interaction to guide transaction-adaptive aggregation. InterFuse encodes each model's representation, prediction score, and input coverage as a model-specific token, using learned states for unavailable outputs. Availability-aware attention updates these tokens and produces a shared fusion state. This state then guides aggregation over available models using their updated representations, original predictions, and input coverage. We evaluate InterFuse on an industrial payment replay and three public financial benchmarks using the same upstream models across fusion methods. InterFuse achieves the highest top-ranked recall and AUC on all four datasets, improving Recall@1\% by 1.11--8.09 percentage points over the strongest baselines. Ablations show the benefits of cross-model interaction and adaptive aggregation, while industrial analyses find larger interaction gains when model predictions disagree. Additional analyses examine robustness to missing outputs and the trade-off between ranking gains and fusion latency.
\end{abstract}
\ccsdesc[500]{Computing methodologies~Neural networks}
\ccsdesc[300]{Information systems~Learning to rank}
\keywords{Fraud detection, model fusion, representation learning, attention mechanisms}
\maketitle
\raggedbottom

\begin{figure}[!t]
\centering
\includegraphics[width=\columnwidth]{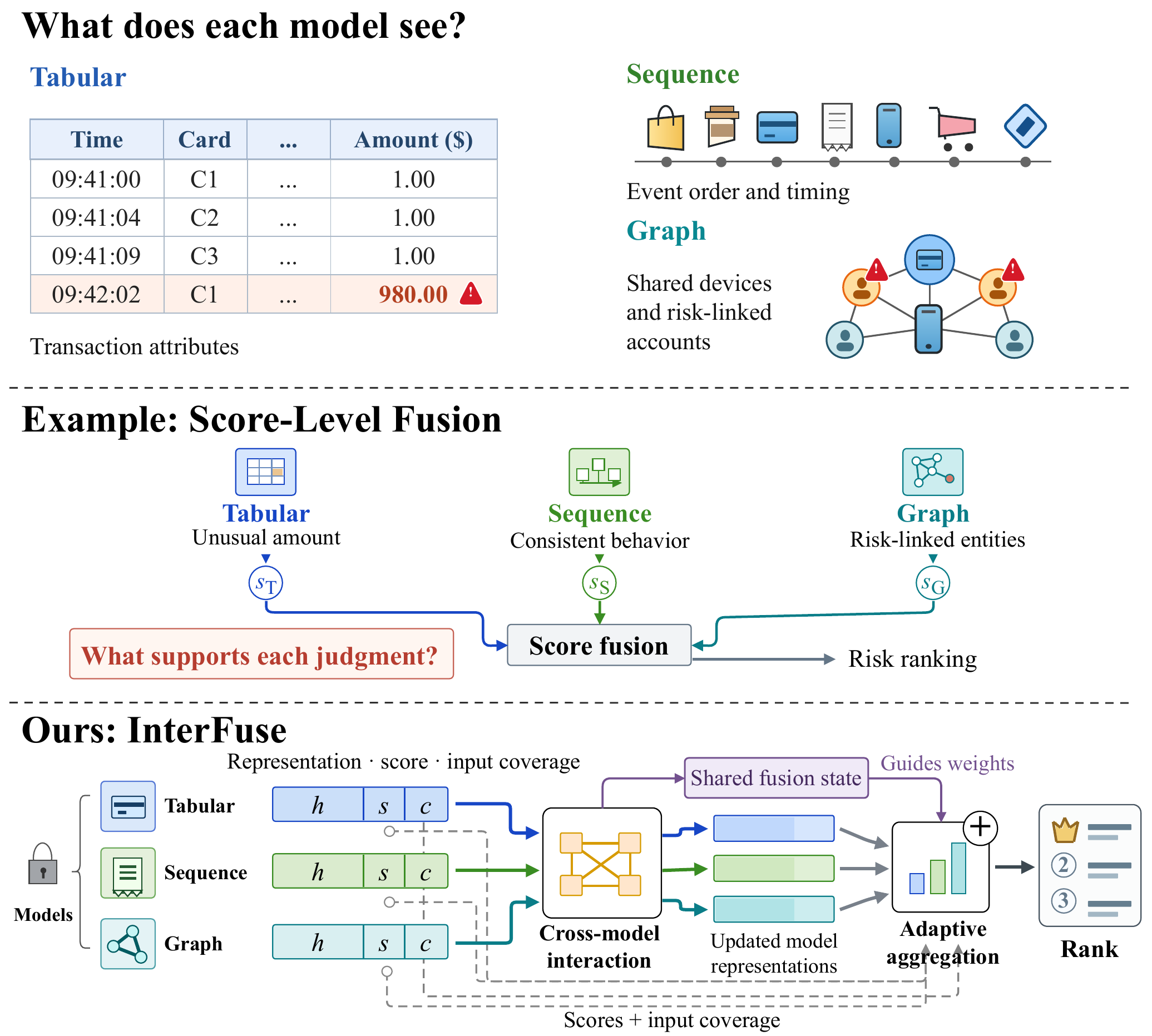}
\caption{Risk views and model fusion. InterFuse uses cross-model interaction to guide adaptive aggregation.}
\Description{Three vertically stacked sections. At the top, an illustrative transaction table on the left shows repeated one-dollar charges followed by a 980-dollar payment. A sequence on the upper right shows event order and timing; a graph below it shows shared devices and risk-linked accounts. The middle section illustrates score-level fusion of tabular, sequence, and graph predictions. The bottom section shows InterFuse updating model representations through cross-model interaction and producing a shared fusion state. Updated representations enter adaptive aggregation, while the shared fusion state guides the aggregation weights. Dashed paths preserve direct access to original scores and input coverage. The table values are illustrative, not experimental observations.}
\label{fig:example}
\end{figure}

\section{Introduction}
Web payment platforms handle massive volumes of transactions but can manually review only a small fraction of them. Identifying high-risk transactions therefore requires integrating information from multiple sources. Transaction attributes characterize the current payment, behavioral histories capture patterns in user, card, and device activity~\cite{padhi2021}, and graph models reveal connections among accounts and transactions~\cite{lu2022,yang2025}. Recent financial benchmarks also incorporate both transaction behavior and relational information~\cite{chen2026}. These complementary sources reflect different aspects of the same transaction and may yield divergent risk assessments. When modeled separately using tabular, sequential, and graph-based approaches, their outputs must be fused to generate a unified risk ranking.

Existing approaches combine these model outputs at different levels. Score-level fusion integrates individual predictions through fixed weights or learned functions~\cite{lakshminarayanan2017,wolpert1992}. Representation-level fusion additionally leverages the features learned by each model, which may encode information not captured in its prediction score. Beyond simple concatenation, the Collaborative Experts model captures relationships among experts to gate their features~\cite{liu2019}, while MMT uses attention to exchange information across model representations~\cite{gabeur2020}. Other approaches model dependencies within and across modalities~\cite{madaan2024}, learn to combine different types of interactions~\cite{xin2025}, or handle missing inputs~\cite{nguyen2026,zhang2023}. These methods provide different ways to model relationships among information sources and adjust their contributions. For independently trained risk models, an important question is how information learned through such interactions should guide the final aggregation when each model also provides its own prediction and input coverage.

Consider a large payment that appears risky based on its transaction attributes but is consistent with the customer's spending history. A connection to a suspicious account may further alter this assessment. The contribution of each model therefore depends not only on its own output but also on what other models reveal about the transaction. The inputs available to each model also matter. For example, a history model may yield a similar prediction whether it is given only a few past transactions or extensive recent activity, even though the information available to the model differs. Some models may produce no output at all. Figure~\ref{fig:example} illustrates these different sources of information and the task of combining their outputs. These considerations raise two related challenges:
\begin{itemize}
    \item \textbf{Cross-model interaction.} How can model representations be updated using information from other models while accounting for differences in input coverage and availability?
    \item \textbf{Interaction-guided adaptive aggregation.} How can the shared context from cross-model interaction guide each model's contribution while retaining its original prediction and input coverage?
\end{itemize}

To address these challenges, we propose \textbf{InterFuse, an interaction-guided adaptive fusion framework for heterogeneous risk models.} InterFuse represents each model's learned features, prediction score, and input coverage as a model-specific token. Learned identities distinguish the models, while learned missing states encode unavailable outputs. Availability-aware attention enables models to exchange information, yielding updated representations and a shared fusion state that captures cross-model information. This shared state then guides adaptive aggregation across the available models. Each updated representation is combined with its original prediction and coverage, so that aggregation can leverage both cross-model information and each model's own assessment. The pooled representation and the shared fusion state jointly determine the final risk score. InterFuse trains only the fusion network, enabling low-cost reuse of existing business model capabilities.

In summary, the main contributions of our work are as follows:
\begin{itemize}
    \item We propose \textbf{InterFuse}, a framework for combining independently trained heterogeneous risk models. It updates model representations through cross-model interaction and uses the resulting shared context to guide transaction-adaptive aggregation.
    \item We develop availability-aware attention to handle missing outputs and aggregation guided by a shared fusion state. Original predictions and input coverage are used at both stages, allowing the fusion network to combine updated representations with each model's original information.
    \item Experiments on an industrial payment replay and three public benchmarks show improved ranking performance over the strongest baselines. Controlled comparisons show that model-level updates add to global fusion, and that explicit shared-state conditioning further improves adaptive aggregation. Industrial analyses show larger interaction gains when model predictions disagree.
\end{itemize}

\section{Related Work}
\textbf{View-specific models and model reuse.}
Tabular, sequence, and graph models learn from different data structures~\cite{gorishniy2025tabm,padhi2021,lu2022,yang2025}, with financial benchmarks covering transaction- and client-level tasks~\cite{chen2026,mollaev2025}. Existing model-reuse methods connect frozen models through trainable querying or cross-attention~\cite{li2023,bansal2024}, or align heterogeneous encoders for merging and fusion~\cite{hemker2025lego}. InterFuse fuses representations and predictions from fixed view-specific models, accounting for output availability and input coverage.

\textbf{Cross-model interaction and aggregation.}
Fusion methods aggregate model predictions or integrate learned representations. MarkerFusion~\cite{xu2026marker} aggregates detector predictions through a domain-conditioned probabilistic model for unsupervised and semi-supervised threat detection. Representation-level methods use set aggregation~\cite{zaheer2017,lee2019}, expert gating~\cite{liu2019}, and cross-modal attention~\cite{gabeur2020,tsai2019,chen2025xfi}. Shared bottlenecks and latent spaces mediate information exchange~\cite{nagrani2021,hemker2024}, while other designs use joint and individual predictors, interaction experts, or sequential modules~\cite{madaan2024,xin2025,swamy2023}. Complementary encoders and adaptive routing offer further fusion strategies~\cite{shi2025,wu2025m4oe,li2025routing}. InterFuse uses the shared fusion state from model interaction to determine each model's weight for the current transaction. Both interaction and aggregation use the models' original predictions and input coverage.

\textbf{Fusion with incomplete inputs.}
Prior work addresses unseen modality combinations~\cite{zhang2023}, unequal availability and learning imbalance~\cite{nguyen2026,wei2024}, and representation collapse caused by noisy features~\cite{chaudhuri2025}. Other methods reconstruct missing modalities~\cite{dai2025,ke2025,guan2026} or enrich observed ones using other samples~\cite{zhuang2025hme}. These settings concern both absent inputs and unequal contributions from observed sources, with different assumptions about the available representations. InterFuse represents missing outputs with learned states and uses availability to control attention and pooling, while coverage describes available inputs.

\section{Problem Setup}
We study the fusion of independently trained view-specific models for risk ranking. Given an event $x$ with binary label $y\in\{0,1\}$, our goal is to learn a fusion function $f(x)$ that combines the outputs of these models to produce a single risk score. In payment risk assessment, events are transactions, and higher scores indicate greater predicted risk.

We consider $R$ models, whose architectures and representation dimensions may differ. For each model $r$, a binary indicator $a_r(x)\in\{0,1\}$ indicates whether its output is available. An available model provides a learned representation $h_r(x)\in\mathbb{R}^{d_r}$ and a prediction score $s_r(x)$ expressed as a logit. Its input coverage vector $c_r(x)\in\mathbb{R}^{k_r}$ captures properties of the observed input, such as history length, recency, or graph degree. Availability and coverage capture different conditions: a model with a short history can still provide a valid output. We denote the set of available models by $\mathcal{A}(x)=\{r:a_r(x)=1\}$ and assume that at least one model is available for each event.

All model outputs pertain to the same event. Historical, relational, and coverage inputs draw on information available at the prediction time of that event. We train the fusion function using event labels while keeping the upstream models and their original prediction heads frozen. The fusion network learns from the available representations and predictions, together with input coverage and the pattern of model availability. For readability, we omit $x$ from individual model outputs below.

The fusion function produces a scalar logit for ranking. Given a review budget of $q\%$, the platform selects the highest-scoring $q\%$ of events. Recall@$q$\% measures the fraction of all positive events recovered within this selection. We describe the task for a single binary target; for multi-target tasks, the fusion model produces one logit per target and ranks events separately for each target.

\section{InterFuse}
\begin{figure*}[t]
\centering
\includegraphics[width=0.97\textwidth]{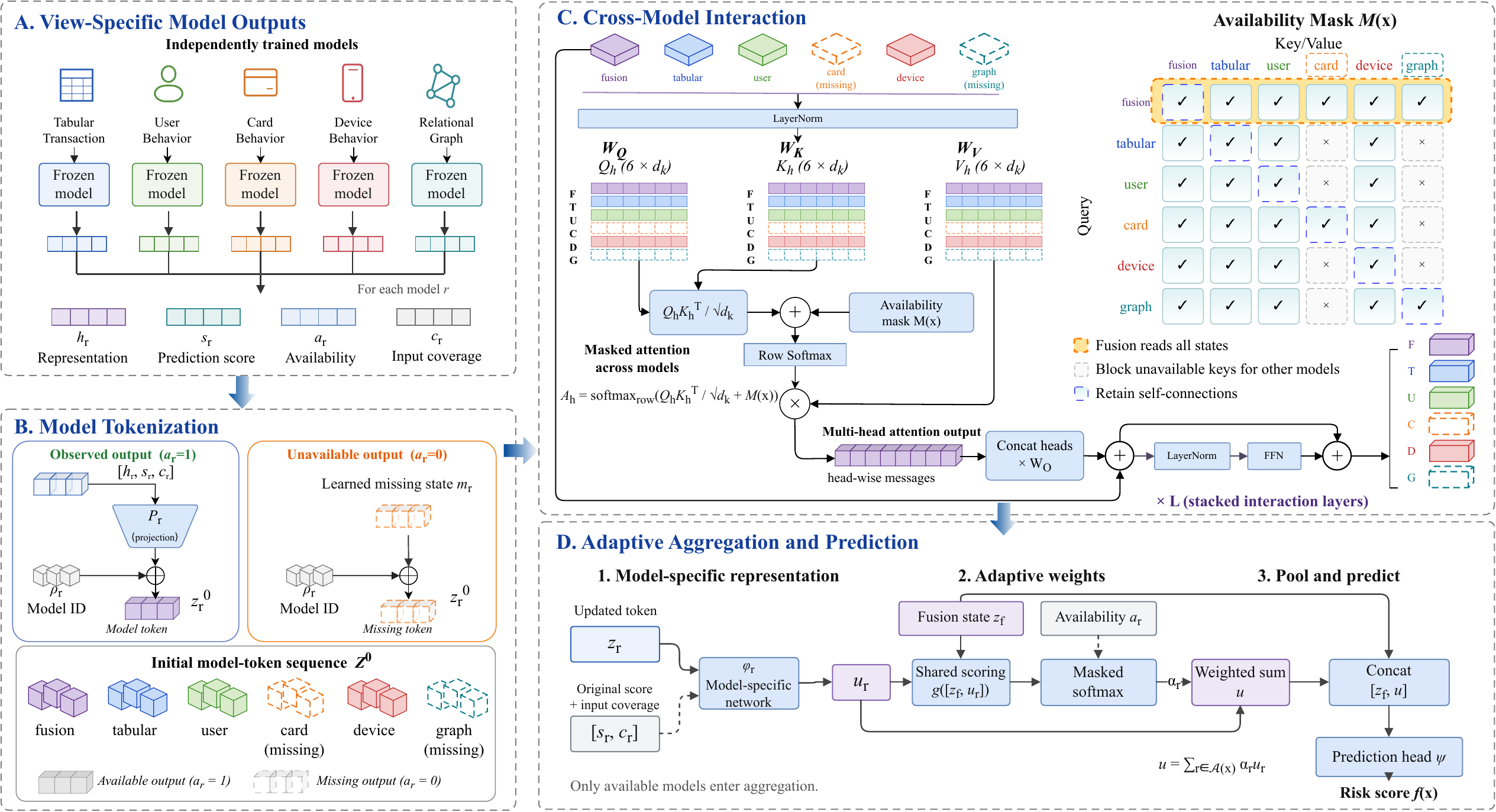}
\caption{InterFuse architecture. Frozen model outputs are encoded as model tokens, updated through availability-aware attention, and adaptively aggregated for risk prediction. Card and graph outputs are unavailable in this example.}
\Description{Four-panel architecture diagram. Five frozen models feed model-identified tokens. Multi-head attention uses an availability mask: the fusion token reads all states, other model tokens cannot read unavailable keys, and self-connections are retained. Available model-specific representations are weighted, summed, and concatenated with the fusion state to predict a risk-ranking score.}
\label{fig:architecture}
\end{figure*}

InterFuse updates model representations through cross-model attention, producing a shared fusion state that guides subsequent aggregation (Figure~\ref{fig:architecture}). Original predictions and input coverage inform both interaction and aggregation, linking cross-model context to each model's contribution to the final risk score.

\subsection{Representing Model Outputs}
For an available model, its model token encodes three kinds of information for joint risk assessment. The representation $h_r\in\mathbb R^{d_r}$ supplies learned features, the score $s_r$ records the model's risk prediction, and $c_r$ describes its input coverage. The fusion network can therefore use both the existing prediction and features that become useful alongside other models.

For an available model, InterFuse concatenates these fields, projects them to a shared width, and adds a learned model identity. Availability selects this projected representation or a learned missing state:
\begin{equation}
z_r^{0}=\rho_r+
\begin{cases}
P_r([h_r\cat s_r\cat c_r]), & a_r=1,\\
m_r, & a_r=0.
\end{cases}
\label{eq:token}
\end{equation}
Here $P_r$ maps the available fields to width $d$, and $\rho_r,m_r\in\mathbb R^d$ are learned embeddings. The projections are model-specific because the input dimensions and feature meanings can differ across sources. Their outputs share the width required by the subsequent attention layers.

Availability distinguishes an absent output from an output based on sparse observations. A sequence model with a short history contributes its representation and coverage through the available branch of Equation~\ref{eq:token}. An unavailable model instead contributes its learned missing state. Adding the model identity in both cases lets the interaction layers distinguish which source supplied an output and which source is absent.

We prepend a learned fusion token and stack the states into $Z^0=[z_f^0;z_1^0;\ldots;z_R^0]\in\mathbb R^{(R+1)\times d}$. The fusion token starts from the same vector for every event and becomes event-specific through attention. Each model has one token regardless of the size of its raw input. Model identity is attached to each model token; the sequence order is used to organize the computation, without an additional positional embedding.

\subsection{Cross-Model Interaction}
Cross-model interaction updates each model's representation using information from other models before aggregation assigns its weight. A sequence-model token can incorporate tabular information about the current payment and graph information about related entities. The same sequence output can therefore produce different updated tokens for different transaction contexts. These updates occur within the fusion network; the view-specific models and their original outputs remain fixed.

The fusion and model tokens pass through $L$ pre-normalized Transformer blocks. Each attention head projects the normalized states into queries, keys, and values. For one head, scaled dot-product attention~\cite{vaswani2017} takes the form
\begin{equation}
\operatorname{Attn}(Z)=\operatorname{softmax}_{\mathrm{row}}\!\left(
\frac{QK^\top}{\sqrt{d_k}}+M(x)\right)V,
\label{eq:attention}
\end{equation}
where $d_k$ is the head width and softmax is applied over key positions in each row. All heads use the same event-dependent mask $M(x)$, but learn separate projections. Their outputs are concatenated and projected back to token width $d$.

\textbf{Communication under missing inputs.}
The fusion state records both available model outputs and the pattern of missing sources. Other model tokens exchange observed information without directly reading another model's missing state. Index the fusion token by $0$ and model tokens by $1,\ldots,R$. The additive mask is
\begin{equation}
M_{ij}(x)=
\begin{cases}
0, & i=0\ \text{or}\ j=0\ \text{or}\ i=j,\\
0, & j\geq 1\ \text{and}\ a_j(x)=1,\\
-\infty, & \text{otherwise}.
\end{cases}
\label{eq:mask}
\end{equation}
The fusion row admits every model state. Model-token rows admit available keys, the fusion key, and self-connections. Blocked keys receive zero attention weight after softmax.

In Figure~\ref{fig:architecture}, the available tabular, user, and device tokens read one another, while the missing card and graph keys are blocked for these queries. The fusion token reads all five model states, including the two missing states. Queries for unavailable models retain self-connections and access to observed states, but are excluded from final pooling. In subsequent layers, the fusion token can pass information about the available source combination back to the model tokens.

\textbf{Updating each token.}
Writing the complete multi-head operation as $\operatorname{MHA}$, one block computes
\begin{equation}
\begin{aligned}
\widetilde Z^\ell &= Z^\ell+
\operatorname{MHA}(\operatorname{LN}(Z^\ell),M),\\
Z^{\ell+1} &= \widetilde Z^\ell+
\operatorname{FFN}(\operatorname{LN}(\widetilde Z^\ell)).
\end{aligned}
\label{eq:block}
\end{equation}
The feed-forward network acts separately on each token, using shared weights across positions within the block. Residual connections retain each state alongside its learned update. We denote the final model states by $z_r$ and the final fusion state by $z_f$.

The updated model representation $z_r$ contains information from other models for the current event. The shared fusion state $z_f$ guides the subsequent aggregation.

\subsection{Interaction-Guided Adaptive Aggregation}
Interaction-guided adaptive aggregation determines how the updated model tokens contribute to the current transaction. Each model's local network processes its updated representation, original prediction, and input coverage. The shared fusion state then guides their relative weights.

\textbf{Model-specific representations and aggregation weights.}
For each available model, a small network $\phi_r$ maps the updated token, original logit, and coverage to $u_r\in\mathbb R^{d_u}$. These networks are model-specific, and their outputs share width $d_u$. The shared scoring network $g$ is a two-layer MLP with a GELU activation and hidden width $d_u$. It jointly transforms the concatenated fusion state and model representation before producing a scalar score. The scores are normalized over available models:
\begin{equation}
\begin{aligned}
u_r &= \phi_r([z_r\cat s_r\cat c_r]),\\
\alpha_r &= \frac{\exp g([z_f\cat u_r])}
{\sum_{j\in\mathcal A(x)}\exp g([z_f\cat u_j])}.
\end{aligned}
\label{eq:pool}
\end{equation}
Normalization includes only available models. The pooled representation is $u=\sum_{r\in\mathcal A(x)}\alpha_r u_r$. The weights sum to one over this set; unavailable models receive zero weight. If only one model is available, its pooling weight is one, while the fusion state still incorporates the pattern of missing sources.

The nonlinear scoring network allows the shared fusion state to influence the relative model weights. A model's contribution can change with its own output, its input coverage, and the information supplied by other models.

The two entry points for scores and coverage serve distinct purposes. In Equation~\ref{eq:token}, they help shape the messages exchanged during interaction. In Equation~\ref{eq:pool}, they provide direct access to each model's original assessment and input conditions after its token has been updated. Aggregation can use this original information alongside the learned cross-model context.

\textbf{Prediction.}
The final representation concatenates the shared fusion state $z_f$ and the weighted model information $u$. A direct prediction head maps their concatenation to a risk logit:
\begin{equation}
f(x)=\psi([z_f\cat u]).
\label{eq:direct}
\end{equation}
Here $\psi$ is the trainable prediction network. The model scores $s_r$ enter through the model tokens and $u_r$, so their role in the final ranking is learned jointly with the fusion layers.

\begin{table*}[t]
\centering
\caption{Overall ranking performance (mean $\pm$ standard deviation over five fusion-training seeds). Industrial reports gains over fixed B1 (pp); public datasets report percentages. B1 is fixed across seeds. R@1 denotes Recall@1\%; MBD averages four targets. Bold and underline mark the best and second-best means.}
\label{tab:main}\fontsize{8.5}{9.9}\selectfont\setlength{\tabcolsep}{1.2pt}
\renewcommand{\arraystretch}{1.36}
\begin{tabular*}{0.965\textwidth}{@{\extracolsep{\fill}}l*{12}{r}@{}}
\toprule
&\multicolumn{3}{c}{Industrial ($\Delta$pp)}&\multicolumn{3}{c}{TransXion (\%)}&\multicolumn{3}{c}{IBM CCT (\%)}&\multicolumn{3}{c}{MBD (\%)}\\
\cmidrule(lr){2-4}\cmidrule(lr){5-7}\cmidrule(lr){8-10}\cmidrule(l){11-13}
Method & AUC & AP & R@1 & AUC & AP & R@1 & AUC & AP & R@1 & AUC & AP & R@1\\
\midrule
B1 & +0.00 & +0.00 & +0.00 & 82.47 & 13.21 & 39.47 & 94.06 & 3.06 & 34.01 & 68.20 & 0.396 & 3.18\\
B5 & +0.85{\scriptsize$\pm$0.11} & +0.37{\scriptsize$\pm$0.35} & +1.77{\scriptsize$\pm$0.21} & 93.69{\scriptsize$\pm$0.07} & 18.85{\scriptsize$\pm$0.41} & 60.31{\scriptsize$\pm$0.52} & \underline{95.33}{\scriptsize$\pm$0.07} & 4.09{\scriptsize$\pm$0.25} & 36.51{\scriptsize$\pm$0.46} & 57.62{\scriptsize$\pm$0.42} & 0.308{\scriptsize$\pm$0.009} & 2.63{\scriptsize$\pm$0.17}\\
\midrule
Concat & \underline{+1.78}{\scriptsize$\pm$0.17} & \underline{+2.67}{\scriptsize$\pm$0.22} & +1.99{\scriptsize$\pm$0.44} & 94.30{\scriptsize$\pm$0.07} & 18.76{\scriptsize$\pm$0.37} & 62.64{\scriptsize$\pm$0.62} & 94.69{\scriptsize$\pm$0.40} & 4.69{\scriptsize$\pm$1.07} & 36.25{\scriptsize$\pm$1.75} & 72.84{\scriptsize$\pm$0.81} & \underline{0.647}{\scriptsize$\pm$0.024} & 6.20{\scriptsize$\pm$0.14}\\
Sum-pooling & -0.67{\scriptsize$\pm$0.13} & +1.37{\scriptsize$\pm$0.31} & +1.81{\scriptsize$\pm$0.21} & 94.60{\scriptsize$\pm$0.08} & 19.26{\scriptsize$\pm$0.65} & 62.97{\scriptsize$\pm$0.96} & 94.62{\scriptsize$\pm$0.13} & 4.54{\scriptsize$\pm$0.28} & 36.07{\scriptsize$\pm$0.35} & 71.55{\scriptsize$\pm$0.73} & 0.605{\scriptsize$\pm$0.032} & 5.97{\scriptsize$\pm$0.24}\\
\midrule
CE & -0.53{\scriptsize$\pm$0.14} & +1.19{\scriptsize$\pm$0.48} & +1.71{\scriptsize$\pm$0.71} & 94.83{\scriptsize$\pm$0.06} & 18.89{\scriptsize$\pm$0.82} & 62.75{\scriptsize$\pm$1.25} & 94.85{\scriptsize$\pm$0.30} & 4.74{\scriptsize$\pm$0.61} & 36.84{\scriptsize$\pm$1.39} & 72.38{\scriptsize$\pm$0.39} & 0.530{\scriptsize$\pm$0.013} & 4.48{\scriptsize$\pm$0.43}\\
MMT & -0.94{\scriptsize$\pm$0.25} & +1.67{\scriptsize$\pm$0.87} & +1.96{\scriptsize$\pm$1.30} & 94.63{\scriptsize$\pm$0.28} & 17.90{\scriptsize$\pm$1.00} & 61.64{\scriptsize$\pm$1.51} & 95.29{\scriptsize$\pm$0.66} & 5.11{\scriptsize$\pm$1.06} & 39.34{\scriptsize$\pm$1.82} & 72.25{\scriptsize$\pm$0.36} & 0.533{\scriptsize$\pm$0.033} & 5.32{\scriptsize$\pm$0.52}\\
UMI & +0.43{\scriptsize$\pm$0.11} & -6.35{\scriptsize$\pm$0.23} & \underline{+3.72}{\scriptsize$\pm$0.27} & 98.59{\scriptsize$\pm$0.09} & 28.87{\scriptsize$\pm$0.63} & \underline{88.58}{\scriptsize$\pm$0.98} & 92.93{\scriptsize$\pm$0.21} & \underline{12.16}{\scriptsize$\pm$0.27} & \underline{53.37}{\scriptsize$\pm$0.92} & 72.35{\scriptsize$\pm$0.32} & 0.562{\scriptsize$\pm$0.062} & 5.28{\scriptsize$\pm$0.39}\\
I$^2$M$^2$ & -0.13{\scriptsize$\pm$0.21} & -3.07{\scriptsize$\pm$0.47} & +2.87{\scriptsize$\pm$0.65} & \underline{99.10}{\scriptsize$\pm$0.09} & \underline{29.47}{\scriptsize$\pm$0.34} & 87.14{\scriptsize$\pm$0.67} & 94.92{\scriptsize$\pm$0.30} & 10.88{\scriptsize$\pm$0.39} & 51.27{\scriptsize$\pm$1.36} & \underline{73.24}{\scriptsize$\pm$0.75} & 0.596{\scriptsize$\pm$0.016} & 4.76{\scriptsize$\pm$0.26}\\
MMPareto & +1.67{\scriptsize$\pm$0.17} & +0.37{\scriptsize$\pm$0.15} & +0.50{\scriptsize$\pm$0.34} & 93.91{\scriptsize$\pm$0.13} & 18.96{\scriptsize$\pm$0.47} & 62.20{\scriptsize$\pm$0.74} & 94.58{\scriptsize$\pm$0.18} & 4.78{\scriptsize$\pm$0.71} & 36.69{\scriptsize$\pm$0.79} & 72.76{\scriptsize$\pm$0.47} & 0.596{\scriptsize$\pm$0.036} & 6.36{\scriptsize$\pm$0.54}\\
I$^2$MoE & +1.29{\scriptsize$\pm$0.26} & +2.59{\scriptsize$\pm$0.68} & +1.48{\scriptsize$\pm$0.69} & 94.50{\scriptsize$\pm$0.09} & 18.67{\scriptsize$\pm$0.98} & 62.20{\scriptsize$\pm$0.94} & 94.93{\scriptsize$\pm$0.49} & 5.02{\scriptsize$\pm$0.78} & 37.43{\scriptsize$\pm$2.49} & 59.68{\scriptsize$\pm$0.53} & 0.418{\scriptsize$\pm$0.020} & 2.60{\scriptsize$\pm$0.51}\\
BALM & +1.46{\scriptsize$\pm$0.38} & +0.64{\scriptsize$\pm$0.54} & +1.27{\scriptsize$\pm$0.38} & 94.75{\scriptsize$\pm$0.08} & 18.97{\scriptsize$\pm$0.70} & 62.08{\scriptsize$\pm$0.64} & 94.87{\scriptsize$\pm$0.30} & 5.48{\scriptsize$\pm$0.82} & 38.31{\scriptsize$\pm$1.32} & 72.94{\scriptsize$\pm$0.74} & 0.590{\scriptsize$\pm$0.040} & \underline{6.87}{\scriptsize$\pm$0.67}\\
\midrule
\rowcolor{rtrow}InterFuse & \textbf{+1.95}{\scriptsize$\pm$0.16} & \textbf{+3.10}{\scriptsize$\pm$0.28} & \textbf{+5.44}{\scriptsize$\pm$0.37} & \textbf{99.15}{\scriptsize$\pm$0.18} & \textbf{29.51}{\scriptsize$\pm$0.67} & \textbf{89.69}{\scriptsize$\pm$0.92} & \textbf{96.82}{\scriptsize$\pm$0.23} & \textbf{22.36}{\scriptsize$\pm$0.29} & \textbf{61.46}{\scriptsize$\pm$1.46} & \textbf{77.80}{\scriptsize$\pm$0.90} & \textbf{0.677}{\scriptsize$\pm$0.033} & \textbf{10.20}{\scriptsize$\pm$0.62}\\
\bottomrule
\end{tabular*}

\end{table*}

\subsection{Training and Inference}
\textbf{Fusion training.}
Fusion training learns to integrate existing model outputs. Each event supplies aligned representations, prediction scores, availability indicators, and input coverage. The trainable parameters comprise the model-specific projections, identity and missing embeddings, fusion token, interaction layers, aggregation networks, and prediction head. The upstream encoders and their original prediction heads remain fixed, allowing their outputs to be cached and reused across fusion-training iterations.

Let $\theta$ collect the fusion parameters and $N$ denote the number of fusion-training events. Training minimizes binary cross-entropy evaluated on the output logits:
\begin{equation}
\mathcal L=\frac1N\sum_{i=1}^N
\ell_{\mathrm{BCE}}(f_\theta(x_i),y_i).
\label{eq:loss}
\end{equation}
The loss $\ell_{\mathrm{BCE}}$ uses the class weights of the corresponding training configuration; for multiple targets, it averages the target-wise losses. Optimizer weight decay is specified separately from the prediction loss. Supervision reaches the model-specific projection networks and their pooling weights through the final prediction.

\textbf{Inference and cost.}
For a new transaction, InterFuse constructs observed or missing tokens from the model outputs, builds the availability mask, and applies the interaction and aggregation layers. The output logit determines the review ranking. The same fusion network accepts different available subsets through the mask and pooling normalization. Algorithm~\ref{alg:routetoken} in Appendix~\ref{app:algorithm} summarizes training and inference.

Each interaction layer processes $R$ model slots and one fusion token, with $(R+1)^2$ query--key scores per head. Missing models retain their slots, so the sequence length is fixed for a given model inventory. For the five-model industrial setting, this gives six positions regardless of history length or graph size. Model projections and feed-forward layers act on these tokens, while the upstream encoders process the raw histories and graphs. We measure fusion cost separately from the cost of producing its inputs.

\section{Experiments}
\label{sec:setup}
Our evaluation examines heterogeneous model fusion for risk ranking. We organize the experiments around four questions: \textbf{RQ1}, does InterFuse improve ranking across tasks and review budgets? \textbf{RQ2}, which inputs and fusion components account for the gains? \textbf{RQ3}, how does performance depend on transaction context and model availability? \textbf{RQ4}, what is the additional cost of fusion?

\vspace{6.2pt}
\subsection{Experimental Setup}
\label{sec:experimental-setup}
\textbf{Datasets and view-specific models.}
We use an industrial payment-risk dataset and three public financial benchmarks. Industrial uses tabular, user-history, card-history, device-history, and graph models. TransXion~\cite{chen2026} uses current-event, sender-history, receiver-history, and prior-event graph models for anti-money-laundering prediction. IBM CCT~\cite{padhi2021} uses tabular, user-history, card-history, and prior-event graph models for card-fraud prediction. These tasks test the same fusion design with different sets of view-specific models.

MBD~\cite{mollaev2025} evaluates next-month product purchase prediction at the client-month level. Its view-specific models describe client profiles, transaction activity, and geolocation activity, with historical inputs restricted by cutoffs based on data generation time. Metrics are averaged equally over four binary targets. This task extends the evaluation to another financial prediction setting and a different model inventory. Table~\ref{tab:data} reports dataset sizes and the model sets used for each task; Appendix~\ref{app:experiment-details} details the chronological splits.

\textbf{Compared methods.}
We compare B1 (single-model reference), B5 (score-only fusion), Concat, Sum-pooling, Collaborative Experts (CE)~\cite{liu2019}, Multi-Modal Transformer (MMT)~\cite{gabeur2020}, UMI~\cite{zhang2023}, I$^2$M$^2$~\cite{madaan2024}, MMPareto~\cite{wei2024}, I$^2$MoE~\cite{xin2025}, and BALM~\cite{nguyen2026}. Concat, Sum-pooling, CE, MMT, and InterFuse receive the same representations $h$, prediction scores $s$, availability indicators $a$, and input coverage $c$. \textbf{We preserve each baseline's core fusion structure and learning mechanisms} while adapting it to the shared model outputs and risk-ranking task. In particular, I$^2$MoE retains its interaction experts, weakly supervised objectives, and sample-dependent weighting. Appendix~\ref{app:baseline-adaptations} gives the adaptation details.

\textbf{Training protocol.}
Within each dataset, all fusion methods use event-aligned outputs from the same view-specific model checkpoints. The upstream encoders and their original prediction heads remain frozen throughout fusion training. Only each method's fusion components and final prediction head are optimized. The key structural comparisons use direct prediction heads, with original model scores supplied as input features rather than additive reference logits. We select fusion checkpoints by validation Recall@1\%, breaking ties by AP, AUC, and the earliest epoch. For the main comparison, we train each fusion method with seeds 17, 42, 202, 512, and 1024, selecting a checkpoint separately for each seed. View-specific model outputs, including B1 predictions, remain fixed across these runs. The component, budget, context, removal, and cost analyses retain the selected seed-42 models. All budgets in a sweep reuse the same checkpoint. Each analysis specifies the evaluation cohort used to compute the reported metrics.

\begin{figure}[t]
\centering\includegraphics[width=\columnwidth]{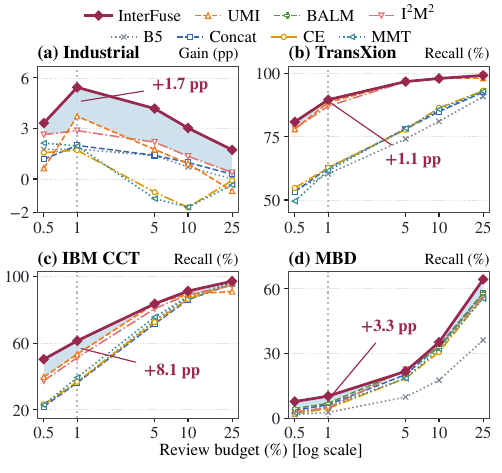}
\Description{A single-column two-by-two layout retains seven method curves per dataset. InterFuse is a solid burgundy line matching Figure 5. Light-blue shading shows where it exceeds the best plotted baseline at each budget; warm shading shows negative gaps. Dotted lines and gain labels mark the one-percent budget. UMI appears on Industrial, TransXion, and IBM CCT, and BALM on MBD.}
\caption{Ranking across review budgets. Shading shows the gap to the best plotted baseline at each budget; labels give 1\% gains. Industrial: pp over B1; others: recall (\%).}
\label{fig:budgets}
\end{figure}

\textbf{Metrics.}
Recall@1\% is the primary metric: the fraction of positive examples recovered in the top-scoring 1\% of the evaluation population. AP and AUC summarize the full ranking. Budget analysis additionally uses 0.5\%, 5\%, 10\%, and 25\%. For a review fraction $b$, Industrial, TransXion, and IBM CCT select $\lfloor bN\rfloor$ events, ordering by descending score and then ascending event ID to break ties. MBD selects $\lceil bN\rceil$ client-months per target, preserves row order for ties, and averages recall equally over four targets. Public results are percentages. Industrial results are percentage-point differences from the same fixed B1 reference at each budget. Table~\ref{tab:main} reports means and standard deviations over the five fusion-training seeds; B1 has no fusion-training variation. Other tables and figures report the selected-model results described above.

\subsection{Overall Ranking Performance (RQ1)}
InterFuse achieves the highest mean Recall@1\% on all four datasets (Table~\ref{tab:main}). It improves over the strongest baseline mean by 1.72 points on Industrial, 1.11 on TransXion, 8.09 on IBM CCT, and 3.33 on MBD. These comparisons use UMI on the three transaction tasks and BALM on MBD. The gains span different financial targets and view-specific model configurations under the same fusion design.

The gains also hold over methods that receive the same model information. On Industrial, InterFuse improves mean Recall@1\% over Concat, CE, and MMT by 3.45--3.73 points. On IBM CCT, it reaches 61.46\%, compared with 36.25--39.34\% for these methods. The improvements over both feature concatenation and interaction-based baselines motivate the component comparisons in Section~\ref{sec:components}, which examine how model interaction and subsequent aggregation contribute to ranking.

InterFuse also achieves the highest mean AUC and AP on all four tasks. The changes in top-ranked recall and full-ranking metrics differ in magnitude: on MBD, the mean Recall@1\% gain over Concat is 4.00 points, accompanied by a 0.030-point mean AP gain. Recall@1\% directly captures the benefit at the review cutoff, complementing the improvements in full-ranking quality.

\begin{table}[!b]
\centering
\caption{Component analysis, Recall@1\%. Industrial: gains over B1 (pp); public datasets: percentages. Full / Both is the complete configuration.}
\label{tab:ablation}\normalsize\setlength{\tabcolsep}{3pt}
\renewcommand{\arraystretch}{1.10}
\begin{tabularx}{.90\columnwidth}{@{}>{\raggedright\arraybackslash}Xrrrr@{}}
\toprule
Variant & Ind. & TX & CCT & MBD\\
\midrule
\rowcolor{rtrow}InterFuse (Full / Both) & \textbf{+5.44} & \textbf{89.69} & \textbf{61.46} & \textbf{10.20}\\
\addlinespace[2pt]
\multicolumn{5}{l}{\textbf{\textit{Input information}}}\\
w/o representation $h$ & +0.15 & 88.14 & 56.94 & 9.33\\
w/o score $s$ & +2.59 & 89.02 & 59.18 & 9.76\\
w/o coverage $c$ & +2.71 & 89.14 & 59.48 & 9.88\\
\addlinespace[2pt]
\multicolumn{5}{l}{\textbf{\textit{Cross-model interaction}}}\\
Self-only routes (C1) & +2.46 & 61.64 & 38.20 & 9.21\\
Bridge-only & +2.69 & 89.14 & 59.88 & 9.78\\
Standard key mask & +4.94 & 89.14 & 60.40 & 10.07\\
\addlinespace[2pt]
\multicolumn{5}{l}{\textbf{\textit{Aggregation weights}}}\\
Uniform & +0.60 & 88.14 & 57.31 & 9.40\\
Static (global) & +1.93 & 88.91 & 58.15 & 9.66\\
w/o $z_f$ in weights & +2.60 & 89.47 & 59.07 & 10.13\\
\addlinespace[2pt]
\multicolumn{5}{l}{\textbf{\textit{Score/coverage entry points}}}\\
Neither & +0.32 & 88.25 & 56.42 & 9.34\\
Token-only & +1.79 & 88.80 & 58.34 & 9.67\\
Aggregation-only & +1.85 & 89.02 & 59.74 & 9.71\\
\bottomrule
\end{tabularx}

\end{table}

The budget curves show where the advantage extends beyond the primary operating point (Figure~\ref{fig:budgets}). InterFuse leads the plotted methods at the 1\% budget on all four tasks, at all five Industrial budgets, and at four of five budgets on IBM CCT and MBD. On the public tasks, some curves cross at wider budgets. The broad lead on Industrial therefore contrasts with budget-dependent method ordering on the public benchmarks.

\begin{figure*}[t]
\centering\includegraphics[width=.8965\textwidth]{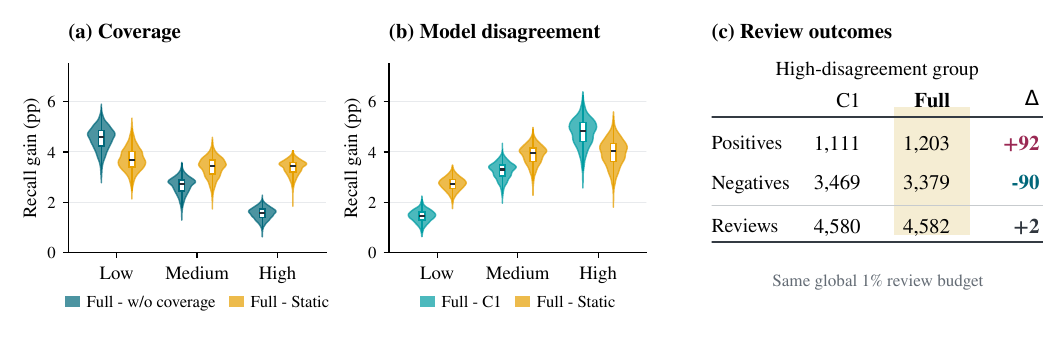}
\Description{Violin and box plots of component recall gains across coverage and prediction-disagreement groups, with a table comparing recovered positives, reviewed negatives, and total reviews in the high-disagreement group.}
\caption{Full's recall gains over (a) w/o coverage and Static across coverage groups and (b) C1 and Static across disagreement groups. Boxes show medians and interquartile ranges; whiskers follow the 1.5-IQR rule. (c) Full versus C1 review outcomes in the high-disagreement group.}
\label{fig:contexts}
\end{figure*}

\subsection{Input and Component Analysis (RQ2)}
\label{sec:components}
The component comparisons examine model interaction, shared-state-guided aggregation, and the information supplied to both stages (Table~\ref{tab:ablation}). Each variant is trained separately under the protocol in Section~\ref{sec:experimental-setup}. Full / Both denotes the complete model. Table~\ref{tab:ablation-extra} reports the corresponding AP and AUC.

\textbf{Cross-model updates improve on global fusion alone.}
C1 retains the fusion token and adaptive aggregation but restricts each model-token query to its own state. Its fusion query still reads all model tokens, so both Full and C1 use information from all available models at prediction time. Full additionally updates individual model states through cross-model interaction before aggregation. This improves Recall@1\% by 2.97 points on Industrial with the same 565.25K fusion parameters, and by 23.26 points on IBM CCT. Updating individual model representations improves ranking even when a shared fusion state already summarizes all model outputs.

Bridge-only adds access to the shared fusion state, creating an indirect communication path across layers. This path recovers most of the gain over C1 on the two public transaction tasks: Bridge-only reaches 89.14\% on TransXion and 59.88\% on IBM CCT, compared with Full's 89.69\% and 61.46\%. On Industrial, Full adds a further 2.74 points over Bridge-only. The benefit of direct model-to-model communication over shared-state bridging varies across tasks.

\begin{samepage}
\textbf{The shared fusion state improves transaction-adaptive weighting.}
Uniform averages available model states, whereas Static learns one global weight per model and renormalizes over available outputs. Static improves on Uniform on all four tasks. Full further improves over Static by 3.51 points on Industrial and 3.31 points on IBM CCT, showing the value of transaction-dependent weights beyond a global model preference.\par
\end{samepage}

The w/o $z_f$ in weights variant is a dynamic-weight control: it uses the same scoring MLP with input $[0\cat u_r]$, so its weights depend on $u_r$ alone. The interaction layers, $u_r=\phi_r([z_r\cat s_r\cat c_r])$, and $z_f$ in the final prediction head remain unchanged. The control therefore retains cross-model information in $u_r$ while removing only the weighting network's explicit shared-state input. Full adds 2.83 points on Industrial and 2.39 points on IBM CCT, with positive gains on TransXion and MBD as well. Explicit shared-state conditioning improves weighting even when the model-specific inputs already contain cross-model information.

\textbf{Original predictions and coverage help at both stages.}
Token-only and Aggregation-only both expose predictions and input coverage to the fusion network, but at different stages. Token-only supplies $(s,c)$ before interaction, Aggregation-only supplies them directly to aggregation, and Neither removes both entries. Full outperforms both single-entry variants on all four tasks; the Industrial gains are 3.65 and 3.58 points, respectively. Direct access to original predictions and input coverage further improves ranking after these fields have already participated in interaction.

\textbf{Input fields and missing-state communication support fusion.}
Removing model representations $h$ lowers Recall@1\% by 5.28 points on Industrial and 4.53 points on IBM CCT, showing the value of learned features beyond original predictions. Scores $s$ and coverage $c$ also improve recall on every task. Full gains 2.73 points over w/o coverage on Industrial, with availability handling retained in both configurations.

Standard key mask blocks unavailable keys for every query, disabling both exceptions in Equation~\ref{eq:mask}: fusion access to missing states and self-connections for missing states. Full exceeds this control by 0.50 points on Industrial and 1.07 points on IBM CCT. This comparison supports the joint missing-state communication rule. Section~\ref{sec:removal} examines its contribution as more outputs become unavailable.

\subsection{Analysis across Transaction Contexts (RQ3)}
Interaction gains are larger when model predictions disagree on Industrial (Figure~\ref{fig:contexts}). Disagreement is the standard deviation of available models' training-referenced percentile scores. All methods share group assignments and independently construct a global top-1\% review queue. We report pointwise 95\% percentile confidence intervals from 2,000 paired user-level bootstrap replicates. Appendix~\ref{app:context-statistics} details the grouping and resampling procedures.

Full's gain over C1 increases from 1.46 points in the low-disagreement group to 4.87 in the high-disagreement group. The high-minus-low contrast is 3.41 points, with confidence interval $[1.92,4.43]$. Full--Static gains also rise, from 2.75 to 4.02 points, with a high-minus-low contrast of 1.27 points and interval $[0.09,2.47]$. Both comparisons link the value of interaction and adaptive aggregation to differences among model predictions.

Under the same global review budget, Full allocates 4,582 reviews to the high-disagreement group and recovers 1,203 positives; C1 allocates 4,580 and recovers 1,111 (Figure~\ref{fig:contexts}c). Within this subgroup, Full therefore finds 92 more positives with two additional reviews. The result connects the interaction gain to improved recovery at nearly identical subgroup review allocations.

Input coverage is most useful when histories are sparse. We measure history coverage by the total valid sequence length across the user-, card-, and device-history models. Full's gain over w/o coverage is 4.61 points in the low-coverage group and 1.58 in the high-coverage group. Their high-minus-low contrast is $-3.03$ points, with interval $[-3.96,-2.03]$. Adaptive aggregation also improves recall across coverage groups, with Full--Static gains of 3.45--3.68 points; their high-minus-low contrast is $-0.23$ with interval $[-1.37,0.51]$.

\begingroup
\setlength{\intextsep}{8pt}
\begin{table}[!b]
\caption{Industrial fusion cost and Recall@1\% gains over B1. Batch-1 latency excludes upstream encoders.}
\label{tab:cost-main}
\centering\normalsize\setlength{\tabcolsep}{3pt}\renewcommand{\arraystretch}{1.10}
\begin{tabularx}{.90\linewidth}{@{}>{\raggedright\arraybackslash}Xrrrr@{}}
\toprule
Method & Params. & p50 & p95 & $\Delta$R@1\\
 & (K) & (ms) & (ms) & (pp)\\
\midrule
Concat & 454.66 & 1.147 & 1.323 & +1.99\\
Sum-pooling & 442.26 & 1.087 & 1.167 & +1.81\\
CE & 410.11 & 11.696 & 11.989 & +1.71\\
MMT & 508.42 & 4.454 & 4.622 & +1.96\\
UMI & 294.43 & 1.314 & 1.334 & +3.72\\
I$^2$M$^2$ & 452.49 & 2.614 & 2.812 & +2.87\\
MMPareto & 343.43 & 1.336 & 2.123 & +0.50\\
I$^2$MoE & 1,077.79 & 5.074 & 6.497 & +1.48\\
BALM & 508.04 & 2.560 & 2.607 & +1.27\\
C1 & 565.25 & 2.808 & 3.277 & +2.46\\
\midrule
\rowcolor{rtrow}InterFuse & 565.25 & 3.044 & 3.570 & \textbf{+5.44}\\
\bottomrule
\end{tabularx}

\end{table}
\endgroup

\subsection{Robustness to Missing Model Outputs (RQ3)}
\label{sec:removal}
We remove user-history, card-history, device-history, and graph outputs separately and jointly, keeping the current-transaction model available. Checkpoints remain fixed. Removal masks $h$, $s$, $c$, and direct aggregation inputs and updates availability. Rates apply to originally available outputs; joint removal independently masks each optional output with probability $p$.

\begin{figure}[!t]
\centering\includegraphics[width=.95\linewidth]{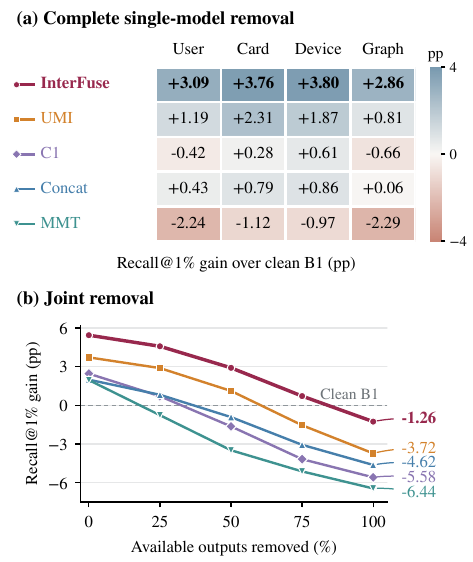}
\Description{A five-method by four-model matrix of complete single-model removal results, followed by five joint-removal curves with labeled endpoints.}
\caption{Industrial output removal. (a) Complete removal of each optional model. (b) Increasing joint removal. Values are Recall@1\% gains over clean B1 (pp).}
\label{fig:removal}
\end{figure}

\textbf{Ranking gains persist across single-model removals.}
InterFuse exceeds clean B1 by 2.86--3.80 points across all four single-model removals and UMI by at least 1.45 points (Figure~\ref{fig:removal}a). These gains are obtained without retraining, showing that the learned fusion function remains useful when the available model set changes. Removing user history or graph has the largest effect, indicating different ranking sensitivities to the missing views. Table~\ref{tab:removal-extra} reports results for all compared methods at intermediate removal rates.

\textbf{The relative advantage persists as joint removal increases.}
InterFuse remains ahead throughout joint removal (Figure~\ref{fig:removal}b), retaining gains over clean B1 of 2.91 points at 50\% and 0.72 at 75\%. The remaining optional outputs still improve ranking over clean B1 at these removal rates. With all optional outputs removed, gains are $-1.26$ points for InterFuse, $-2.18$ for Standard key mask, and $-3.72$ for UMI. The results distinguish two effects: complementary outputs improve absolute recall, while the fusion design matters under the same reduced inputs.

Standard key mask disables fusion-token access to missing states and missing-state self-connections while preserving the rest of the fusion design. Full exceeds it at every reported condition, by 0.63 points at 50\% joint removal. This supports missing-state communication during interaction, beyond excluding unavailable outputs from the final weighting and pooling steps.

\subsection{Computational Cost (RQ4)}

All methods use the same execution settings; latency is measured at batch size 1 over the fusion network only. InterFuse achieves its Industrial ranking gains with 565.25K fusion parameters and 3.044\,ms median fusion latency (Table~\ref{tab:cost-main}). Relative to C1, it improves Recall@1\% by 2.97 points at the same parameter count, with an 8.4\% latency increase over 2.808\,ms. These measurements isolate the extra computation required for fusion of existing model outputs. Attention operates on one token per model and a fusion token, so its sequence length depends on the model inventory rather than raw history length or graph size. The Industrial configuration therefore processes six tokens per transaction.

InterFuse also has higher recall and lower median latency than CE, MMT, and I$^2$MoE. Compared with UMI, it gains 1.71 recall points at a higher median latency; UMI uses 294.43K parameters and 1.314\,ms. The table makes this quality--cost choice explicit across all eleven methods, with upstream encoder computation excluded from the timing boundary. Figure~\ref{fig:cost-quality} in Appendix~\ref{app:cost-quality} plots ranking gains against fusion latency.

\section{Conclusion}
In this work, we studied how to fuse models trained on different transaction views. We proposed InterFuse, which updates model representations through availability-aware attention and uses a shared fusion state to guide adaptive aggregation. Aggregation retains original predictions and input coverage to determine model contributions. Experiments demonstrate consistent improvements in risk ranking across industrial and public benchmarks. Our analyses show that updating individual model representations adds value beyond global summarization. Aggregation further benefits from explicit access to the shared fusion state, original predictions, and input coverage after interaction. Industrial analyses find larger interaction gains when model predictions disagree. Future work could explore how this approach can help integrate existing model capabilities into payment-risk foundation models.

\appendix
\section{Additional Experimental Details}
\label{app:experiment-details}
\subsection{Datasets and Chronological Splits}
\label{app:splits}
\begin{table*}[t]
\caption{Tasks, data splits, and view-specific models. Public Train counts include model and fusion fitting. Positive rates refer to Test; the range for MBD covers its four product-purchase targets.}
\label{tab:data}
\centering\fontsize{9}{11}\selectfont\setlength{\tabcolsep}{4pt}\renewcommand{\arraystretch}{1.25}
\begin{tabular*}{0.96\textwidth}{@{\extracolsep{\fill}}l>{\raggedright\arraybackslash}p{2.1cm}rrr>{\raggedright\arraybackslash}p{1.25cm}>{\raggedright\arraybackslash}p{3.2cm}@{}}
\toprule
Dataset & Task / unit & Train & Validation & Test & Positive rate & View-specific models\\
\midrule
Industrial & Payment risk / event & 5,832,600 & 1,392,465 & 1,366,237 & 0.576\% & Tabular; user, card, device sequences; graph\\
TransXion & AML / event & 1,817,503 & 605,834 & 605,833 & 0.149\% & Tabular; sender, receiver sequences; graph\\
IBM CCT & Card fraud / event & 19,512,454 & 2,436,496 & 2,437,950 & 0.111\% & Tabular; user, card sequences; graph\\
MBD (trx/geo) & Product purchase / client-month & 801,792 & 200,448 & 200,448 & 0.040--0.245\% & Profile; transaction and geolocation summaries\\
\bottomrule
\end{tabular*}
\end{table*}

\textbf{Transaction tasks.}
Industrial, TransXion, and IBM CCT evaluate event-level risk ranking. Their Test sets contain 7,872, 902, and 2,717 positive events, respectively; the corresponding Validation counts are 7,811, 779, and 2,116. TransXion and IBM CCT contain 2,960 and 24,924 training positives. Table~\ref{tab:data} summarizes the tasks, sample counts, positive rates, and model sets.

\textbf{Temporal inputs and model fitting.}
The transaction benchmarks use chronological splits, reflecting the deployment conditions emphasized by TabReD~\cite{rubachev2025}. Sequence and graph inputs precede the scored event. Public datasets use separate partitions for model and fusion fitting; Validation selects checkpoints for Test evaluation. All fusion methods use the same event-level outputs from encoders and prediction heads that remain fixed throughout fusion training.

\textbf{MBD prediction task.}
The trx/geo version predicts next-month purchases for four client-month targets using cutoffs based on data generation time, with equally averaged metrics. Positive counts are 4,267, 386, 3,494, and 2,095 for Train; 644, 110, 576, and 473 for Validation; and 451, 81, 460, and 492 for Test. These targets describe product purchases, extending the evaluation beyond fraud detection.

MLPs encode profile features and fixed summaries of strictly prior transaction and geolocation events into 64-dimensional representations. Model fitting uses the first four training months; fusion fitting uses the next four, with three epochs per stage. Validation covers October--November 2022; Test covers December 2022--January 2023. Each target uses a separate head and a fixed checkpoint across all review budgets.

\subsection{Baseline Adaptations}
\label{app:baseline-adaptations}
\textbf{Common prediction setting.}
B1 is the single-model reference; B5 performs score fusion. Structural comparisons receive aligned representations, scores, availability, and coverage, using model-specific projections for different input dimensions and direct prediction heads. Checkpoint selection follows Section~\ref{sec:experimental-setup}.

\textbf{Simple feature fusion.}
Concat joins projected model outputs; Sum-pooling sums available projections. Both are simple alternatives to cross-model interaction and adaptive aggregation.

\textbf{Collaborative Experts and MMT.}
CE~\cite{liu2019} uses available-pair averaging, shared relation networks, local-plus-relation gates, and a fixed expert-update order. MMT~\cite{gabeur2020} uses aggregate and feature tokens for each model summary, learned model identities and position embeddings, and two post-normalization Transformer layers with four heads. Unavailable keys are masked and their output embeddings zeroed. Prediction retains expert aggregate embeddings, without InterFuse's adaptive pooling.

Both methods classify concatenated normalized expert outputs and availability indicators. Binary heads and cross-entropy replace the retrieval components, without video or language pretraining. MBD uses four heads. MMT initializes each aggregate token from the exported model summary vector.

\textbf{UMI and interaction-based alternatives.}
UMI~\cite{zhang2023} sums available outputs projected into a common space; a second branch learns from reliable unimodal pseudo-labels. Model-specific projections replace token reorganization. I$^2$M$^2$~\cite{madaan2024} integrates joint and available individual predictors, extending its two-modality factorization to the model inventory. I$^2$MoE~\cite{xin2025} uses individual-model, model-pair, and complete-set experts for different interaction types.

\textbf{Training-oriented alternatives.}
MMPareto~\cite{wei2024} balances joint and model-specific gradients. BALM~\cite{nguyen2026} uses feature calibration and gradient rebalancing for unequal availability. Both act on the fusion network, keeping upstream encoders fixed.

\subsection{Transaction Context Groups}
\label{app:context-statistics}
\textbf{History coverage.}
The user-, card-, and device-history models each use events from the 90 days preceding the current transaction, retaining the most recent 300 events when this limit is exceeded. History coverage is the sum of the three valid sequence lengths, excluding padding. Empty histories contribute zero. An event appearing in multiple histories is counted separately because it contributes to each model's input. The aggregate ranges from 0 to 900. Low, Medium, and High coverage correspond to fewer than 50, 50--300, and more than 300 history entries, respectively.

\textbf{Prediction disagreement.}
We use the original prediction scores from the tabular, user-history, card-history, device-history, and graph models. For each model, we construct an empirical score distribution from its available outputs on the training reference set. Each score is mapped to its percentile in this distribution, placing predictions on a common scale of relative risk. For a transaction $x$, disagreement is
\begin{equation}
\begin{aligned}
D(x)&=\sqrt{\frac{1}{m_x}\sum_{r\in\mathcal A(x)}\bigl(q_r(x)-\bar q(x)\bigr)^2},\\
\bar q(x)&=\frac{1}{m_x}\sum_{r\in\mathcal A(x)}q_r(x),
\end{aligned}
\label{eq:disagreement}
\end{equation}
where $\mathcal A(x)$ is the set of available models, $m_x=|\mathcal A(x)|$, and $q_r(x)$ is model $r$'s percentile score. Unavailable outputs are excluded.

We determine cutoffs $d_1$ and $d_2$ from the one-third and two-thirds quantiles of training-reference disagreement values, using transactions with at least two available models. The groups are Low ($D\leq d_1$), Medium ($d_1<D\leq d_2$), and High ($D>d_2$). The reference distributions and cutoffs remain fixed during evaluation and bootstrap resampling. Equal disagreement values remain in the same group. With fewer than two available models, a transaction stays in the global ranking but has no disagreement group.

\textbf{Shared grouping and review allocation.}
All fusion methods use the same groups, defined from the view-specific models' original inputs and predictions. The Industrial cohort contains 1,366,237 events and 7,872 positives. Each method independently selects its global top-1\% queue of 13,662 events; no separate budget is assigned to individual groups. Group recall is the fraction of a group's positive events recovered in the global review queue.

\textbf{Bootstrap confidence intervals.}
We use 2,000 paired bootstrap replicates, sampling users with replacement and retaining all transactions associated with each sampled user. Each replicate uses the same sampled records and multiplicities across methods and groups. Model checkpoints, reference score distributions, and group cutoffs remain fixed. We reconstruct each method's global top-1\% review queue within each replicate and compute group-specific recall gains. For each comparison, the high-minus-low contrast is the gain in the High group minus that in the Low group. We report two-sided, pointwise 95\% percentile confidence intervals using the 2.5th and 97.5th percentiles of the bootstrap contrasts. These intervals describe evaluation-sample uncertainty for the fixed models.

\clearpage
\twocolumn[{
\subsection{Additional Metrics for the Component Analysis}
\label{app:component-metrics}
\begin{minipage}{\textwidth}
\captionsetup{type=table}
\caption{Additional AP and AUC for the component variants in Table~\ref{tab:ablation}. Industrial reports gains over B1 (pp); public datasets report percentages. Each value is a point estimate from the selected model for that variant.}
\label{tab:ablation-extra}
\centering\small\setlength{\tabcolsep}{3pt}\renewcommand{\arraystretch}{1.05}
\begin{tabular*}{0.88\textwidth}{@{\extracolsep{\fill}}l*{8}{r}@{}}
\toprule
&\multicolumn{2}{c}{Industrial ($\Delta$pp)}&\multicolumn{2}{c}{TransXion (\%)}&\multicolumn{2}{c}{IBM CCT (\%)}&\multicolumn{2}{c}{MBD (\%)}\\
Variant & AUC & AP & AUC & AP & AUC & AP & AUC & AP\\
\midrule
InterFuse (Full / Both) & +1.95 & +3.10 & 99.15 & 28.71 & 96.82 & 22.36 & 77.80 & 0.677\\
\addlinespace[2pt]
\multicolumn{9}{l}{\textbf{\textit{Input information}}}\\
w/o representation $h$ & +1.75 & -1.96 & 99.06 & 25.99 & 96.63 & 20.92 & 77.03 & 0.621\\
w/o score $s$ & +1.86 & +0.52 & 99.11 & 27.33 & 96.74 & 21.60 & 77.44 & 0.649\\
w/o coverage $c$ & +1.87 & +1.06 & 99.11 & 27.65 & 96.75 & 21.74 & 77.45 & 0.654\\
\addlinespace[2pt]
\multicolumn{9}{l}{\textbf{\textit{Cross-model interaction}}}\\
Self-only routes (C1) & +1.72 & +2.67 & 94.02 & 18.32 & 95.04 & 5.23 & 76.72 & 0.609\\
Bridge-only & +1.86 & +0.80 & 99.11 & 27.48 & 96.74 & 21.69 & 77.47 & 0.652\\
Standard key mask & +1.86 & +2.85 & 99.14 & 28.66 & 96.67 & 21.26 & 77.35 & 0.612\\
\addlinespace[2pt]
\multicolumn{9}{l}{\textbf{\textit{Aggregation weights}}}\\
Uniform & +1.76 & -1.62 & 99.07 & 26.18 & 96.64 & 21.01 & 77.08 & 0.624\\
Static (global) & +1.83 & +0.03 & 99.10 & 27.12 & 96.71 & 21.49 & 77.28 & 0.643\\
w/o $z_f$ in weights & +1.91 & +2.44 & 99.12 & 28.31 & 96.78 & 22.27 & 77.60 & 0.666\\
\addlinespace[2pt]
\multicolumn{9}{l}{\textbf{\textit{Score/coverage entry points}}}\\
Neither & +1.74 & -2.08 & 99.06 & 25.96 & 96.62 & 20.88 & 77.01 & 0.620\\
Token-only & +1.83 & +0.01 & 99.10 & 27.12 & 96.72 & 21.49 & 77.37 & 0.644\\
Aggregation-only & +1.84 & +0.18 & 99.10 & 27.18 & 96.72 & 21.52 & 77.39 & 0.645\\
\bottomrule
\end{tabular*}

\end{minipage}
\vspace{10pt}
}]

\subsection{Fusion Performance across Model Subsets}
\label{app:model-subsets}
\begin{minipage}{\columnwidth}
\centering
\includegraphics[width=0.95\columnwidth]{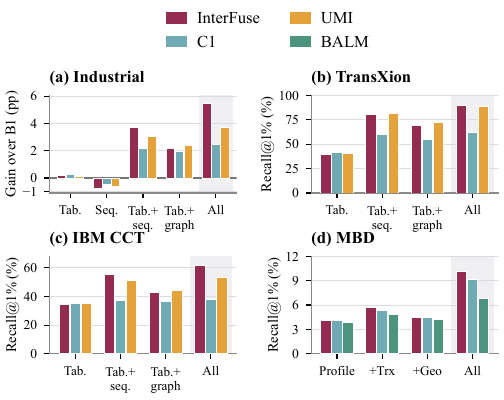}
\captionsetup{type=figure}
\caption{Fusion performance across model subsets. Training and evaluation use the same subset; shading marks all models. Tab.: tabular; Seq.: sequences; +Trx/+Geo: profile plus transaction/geolocation summaries.}
\label{fig:model-subsets}
\end{minipage}

Figure~\ref{fig:model-subsets} uses a separately trained fusion model for each subset, with upstream encoders fixed. Unlike Figure~\ref{fig:removal}, the subset is unchanged between training and evaluation.

The complete model set gives InterFuse its highest recall on each dataset and a larger advantage over C1 than either two-source subset. Performance varies with the model subset: UMI achieves higher recall with tabular and graph models on all three transaction datasets.

\newpage
\subsection{Training and Inference Algorithm}
\label{app:algorithm}
\begingroup
\makeatletter\@twocolumnfalse\makeatother
\SetAlFnt{\fontsize{8}{9.2}\selectfont}
\SetAlCapFnt{\small}
\SetAlCapNameFnt{\small\bfseries}
\SetAlgoNlRelativeSize{-2}
\SetAlgoInsideSkip{smallskip}
\SetAlCapSkip{3pt}
\DontPrintSemicolon
\SetKwInput{KwInput}{Input}
\SetKwInput{KwOutput}{Output}
\SetKwProg{Fn}{function}{:}{end}
\SetKwFunction{Forward}{Forward}
\begin{algorithm}[H]
\caption{Training and Inference of InterFuse}
\label{alg:routetoken}
\KwInput{Aligned model outputs $\mathcal E(x)=\{(h_r,s_r,a_r,c_r)\}_{r=1}^{R}$; fusion-training labels; validation data; depth $L$.}
\KwOutput{Selected fusion parameters $\theta^\star$ and ranking logits.}
\tcp*[l]{Shared forward computation (Eqs.~\ref{eq:token}--\ref{eq:direct})}
\Fn{\Forward{$\mathcal E(x);\theta$}}{
 $\mathcal A\leftarrow\{r:a_r=1\}$ \quad (assumed nonempty)\;
 $z_r^0\leftarrow\rho_r+P_r([h_r\cat s_r\cat c_r]),\ r\in\mathcal A$\;
 $z_r^0\leftarrow\rho_r+m_r,\ r\notin\mathcal A$\;
 $Z^0\leftarrow[z_f^0;z_1^0;\ldots;z_R^0]$\;
 Build $M(x)$ by Eq.~\ref{eq:mask}, retaining fusion and self access\;
 \For{$\ell=0,\ldots,L-1$}{
 $Z^{\ell+1}\leftarrow$ pre-normalized interaction block applied to $(Z^\ell,M(x))$ (Eq.~\ref{eq:block})\;
 }
 $z_f,z_1,\ldots,z_R\leftarrow$ token states in $Z^L$\;
 $u_r\leftarrow\phi_r([z_r\cat s_r\cat c_r]),\ r\in\mathcal A$\;
 $\alpha_{\mathcal A}\leftarrow\operatorname{softmax}_{r\in\mathcal A}\bigl(g([z_f\cat u_r])\bigr)$\;
 $u\leftarrow\sum_{r\in\mathcal A}\alpha_r u_r$\;
 \KwRet{$\psi([z_f\cat u])$}\;
}
\tcp*[l]{Fusion training; view-specific models stay fixed}
Initialize fusion parameters $\theta$\;
\ForEach{training epoch}{
 \ForEach{minibatch $\mathcal B$}{
 $f_\theta(x)\leftarrow\Forward(\mathcal E(x);\theta),\ x\in\mathcal B$\;
 $\mathcal L_{\mathcal B}\leftarrow\frac{1}{|\mathcal B|}\sum_{x\in\mathcal B}\ell_{\mathrm{BCE}}(f_\theta(x),y_x)$\;
 Update only $\theta$ using $\mathcal L_{\mathcal B}$ (Eq.~\ref{eq:loss})\;
 }
 Save the epoch checkpoint and validation metrics\;
}
Select $\theta^\star$ using the validation rule in Sec.~\ref{sec:experimental-setup}\;
\tcp*[l]{Inference; one fusion forward pass}
\KwRet{$\Forward(\mathcal E(x);\theta^\star)$ as the ranking logit}\;
\end{algorithm}
\endgroup

\clearpage
\twocolumn[{
\subsection{Complete Industrial Output-Removal Results}
\label{app:output-removal}
\begin{minipage}{\textwidth}
\captionsetup{type=table}
\caption{Complete Industrial output-removal results. Entries are Recall@1\% gains over the same clean B1 reference (pp). Removal rates apply to originally available outputs. Mask-level dispersion is not reported.}
\label{tab:removal-extra}
\centering\small\setlength{\tabcolsep}{4pt}\renewcommand{\arraystretch}{1.00}
\begin{tabular*}{0.76\textwidth}{@{\extracolsep{\fill}}llrrrrr@{}}
\toprule
View & Method & 0\% & 25\% & 50\% & 75\% & 100\%\\
\midrule
User-history & InterFuse & +5.44 & +5.16 & +4.47 & +3.55 & +3.09\\
 & Standard key mask & +4.94 & +4.63 & +3.71 & +2.88 & +2.05\\
 & C1 & +2.46 & +2.08 & +1.24 & +0.31 & -0.42\\
 & Concat & +1.99 & +1.43 & +0.89 & +0.57 & +0.43\\
 & MMT & +1.96 & +1.61 & +0.53 & -1.12 & -2.24\\
 & UMI & +3.72 & +3.39 & +2.82 & +1.96 & +1.19\\
\addlinespace[2pt]
Card-history & InterFuse & +5.44 & +4.73 & +4.20 & +3.91 & +3.76\\
 & Standard key mask & +4.94 & +4.26 & +3.58 & +3.12 & +2.53\\
 & C1 & +2.46 & +1.63 & +1.07 & +0.52 & +0.28\\
 & Concat & +1.99 & +1.88 & +1.56 & +1.03 & +0.79\\
 & MMT & +1.96 & +0.91 & +0.14 & -0.64 & -1.12\\
 & UMI & +3.72 & +3.11 & +2.67 & +2.45 & +2.31\\
\addlinespace[2pt]
Device-history & InterFuse & +5.44 & +5.32 & +4.98 & +4.29 & +3.80\\
 & Standard key mask & +4.94 & +4.72 & +4.31 & +3.61 & +2.85\\
 & C1 & +2.46 & +2.31 & +1.82 & +1.04 & +0.61\\
 & Concat & +1.99 & +1.60 & +1.26 & +1.03 & +0.86\\
 & MMT & +1.96 & +1.69 & +1.12 & -0.18 & -0.97\\
 & UMI & +3.72 & +3.66 & +3.18 & +2.41 & +1.87\\
\addlinespace[2pt]
Graph & InterFuse & +5.44 & +4.39 & +3.69 & +3.17 & +2.86\\
 & Standard key mask & +4.94 & +3.91 & +3.06 & +2.34 & +1.69\\
 & C1 & +2.46 & +1.52 & +0.86 & +0.15 & -0.66\\
 & Concat & +1.99 & +1.77 & +1.16 & +0.41 & +0.06\\
 & MMT & +1.96 & +0.64 & -0.83 & -1.76 & -2.29\\
 & UMI & +3.72 & +2.66 & +1.77 & +1.18 & +0.81\\
\addlinespace[2pt]
Joint & InterFuse & +5.44 & +4.58 & +2.91 & +0.72 & -1.26\\
 & Standard key mask & +4.94 & +3.81 & +2.28 & -0.41 & -2.18\\
 & C1 & +2.46 & +0.71 & -1.63 & -4.17 & -5.58\\
 & Concat & +1.99 & +0.82 & -0.91 & -3.06 & -4.62\\
 & MMT & +1.96 & -0.76 & -3.48 & -5.12 & -6.44\\
 & UMI & +3.72 & +2.89 & +1.12 & -1.54 & -3.72\\
\addlinespace[2pt]
\bottomrule
\end{tabular*}

\end{minipage}
\vspace{10pt}
}]

Table~\ref{tab:removal-extra} adds intermediate removal rates to Figure~\ref{fig:removal}. Initial availability is 68.72\% for user history, 58.40\% for card history, 62.03\% for device history, and 44.69\% for graph outputs. A 100\% rate removes all originally available outputs of the named model.

The fusion checkpoint remains fixed. Removal updates availability and suppresses the corresponding representations, scores, coverage, and direct aggregation inputs. The current-transaction model remains present. Single-model scenarios affect only the named model; in Joint, each originally available optional output is removed independently with probability $p$.

InterFuse has the highest recall at every reported condition. At complete joint removal, its gain over clean B1 is $-1.26$ points, versus $-2.18$ for Standard key mask and $-3.72$ for UMI.

\subsection{Figure Display Details}
Figure~\ref{fig:contexts}(a--b) displays bootstrap recall-gain distributions (pp). These distributions reflect evaluation-sample variability with model checkpoints held fixed.

\newpage
\subsection{Ranking Quality and Fusion Latency}
\label{app:cost-quality}
\begin{minipage}{\columnwidth}
\centering
\includegraphics[width=0.80\columnwidth]{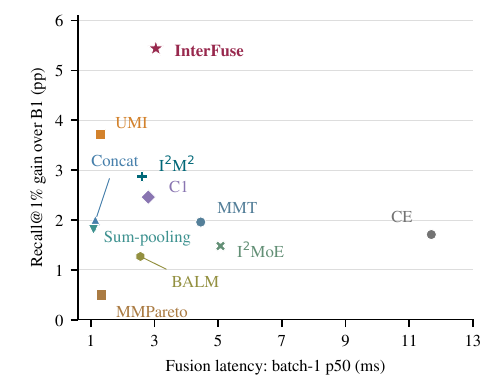}
\captionsetup{type=figure}
\caption{Industrial recall gain versus batch-1 median fusion latency. All eleven methods from Table~\ref{tab:cost-main}; encoders excluded.}
\label{fig:cost-quality}
\end{minipage}

\end{document}